\documentclass[sigconf, screen]{acmart}
\AtBeginDocument{%
  }

\setcopyright{acmlicensed}
\copyrightyear{2025}
\acmYear{2025}
\acmDOI{XXXXXXX.XXXXXXX}
\acmConference[ACM MM]{}{Octobor 27--31,
  2025}{Dublin, Ireland}
\acmISBN{978-1-4503-XXXX-X/2018/06}

\usepackage{multirow}
\usepackage{graphicx}
\usepackage{pifont}
\usepackage{graphicx}
\usepackage{multirow}
\usepackage{multicol}

\usepackage{algorithm}
\usepackage{algorithmic}
\usepackage{amsmath}
\usepackage{caption}
\usepackage{float}
\usepackage{balance}

\usepackage{amsmath}

\begin{document}

\title{Character-Centric Understanding of Animated Movies}

\author{Zhongrui Gui}
\affiliation{%
  \institution{Visual Geometry Group \\ Dept.\ of Engineering Science \\ University of Oxford, UK}
  \country{}}
\email{zhongrui@robots.ox.ac.uk}

\author{Junyu Xie}
\affiliation{%
  \institution{Visual Geometry Group \\ Dept.\ of Engineering Science \\ University of Oxford, UK}
  \country{}}
\email{jyx@robots.ox.ac.uk}

\author{Tengda Han}
\affiliation{%
  \institution{Visual Geometry Group \\ Dept.\ of Engineering Science \\ University of Oxford, UK}
  \country{}}
\email{htd@robots.ox.ac.uk}

\author{Weidi Xie}
\affiliation{%
  \institution{School of Artifitial Intelligence\\Shanghai Jiao Tong University}
  \city{Shanghai}
  \country{China}}
\email{weidi@sjtu.edu.cn}

\author{Andrew Zisserman}
\affiliation{%
  \institution{Visual Geometry Group \\ Dept.\ of Engineering Science \\ University of Oxford, UK}
  \country{}}
\email{az@robots.ox.ac.uk}

\renewcommand{\shortauthors}{Zhongrui Gui, Junyu Xie, Tengda Han, Weidi Xie, and Andrew Zisserman}

\begin{abstract}
Animated movies are captivating for their unique character designs and imaginative storytelling, yet they pose significant challenges for existing recognition systems. Unlike the consistent visual patterns detected by conventional face recognition methods, animated characters exhibit extreme diversity in their appearance, motion, and deformation. In this work, we propose an audio-visual pipeline to enable automatic and robust animated character recognition, and thereby enhance character-centric understanding of animated movies. Central to our approach is the automatic construction of an audio-visual character bank from online sources. This bank contains both visual exemplars and voice (audio) samples for each character, enabling subsequent multi-modal character recognition despite long-tailed appearance distributions.
Building on accurate character recognition, we explore two downstream applications: Audio Description (AD) generation for visually impaired audiences, and character-aware subtitling for the hearing impaired. To support research in this domain, we introduce CMD-AM, a new dataset of $75$ animated movies with comprehensive annotations. Our character-centric pipeline demonstrates significant improvements in both accessibility and narrative comprehension for animated content over prior face-detection-based approaches. For the code and dataset, visit \url{https://www.robots.ox.ac.uk/~vgg/research/animated_ad/}.
\end{abstract}




\begin{teaserfigure}
  \centering
  \includegraphics[width=0.975\textwidth]{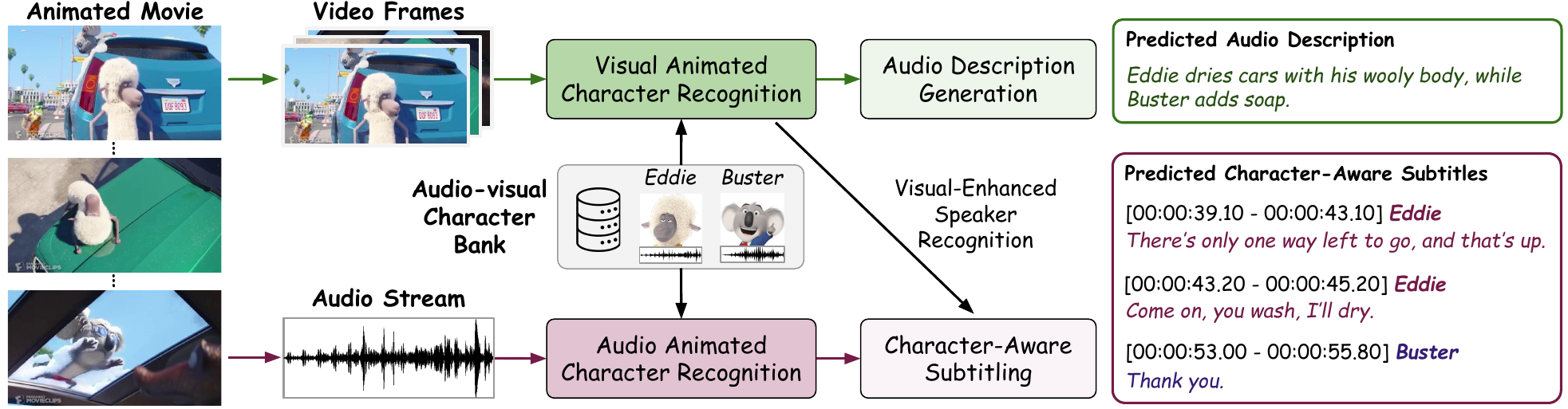}
  \caption{\textit{Character-Centric Understanding of Animated Movies.} We propose a pipeline for recognising characters in animated movies across both visual and audio domains. Our approach involves constructing an audio-visual character bank from online sources, which is then used for video and audio (voice) animated character recognition. We use character recognition for two downstream tasks: Audio Description (AD) generation and character-aware subtitling. The movie clip is from \textit{Sing}.}
  \label{fig:teaser}
\end{teaserfigure}


\maketitle
\section{Introduction}

When we were children, animated movies were more than just entertainment -- they introduced us to our favourite characters, sparked our imagination with captivating stories, 
and left us with lasting memories.
Today, modern computer vision systems are rapidly advancing towards superhuman capabilities across many domains, from recognising bird species to describing fine-grained actions. 
Yet, intriguingly, existing technologies often struggle to comprehend animated movies -- a medium effortlessly understood by young minds. 

Ask any four-year-old child to describe an animation they have just seen, and the reason for this computer vision system failing is obvious: the child names the characters and can recognise them, but current computer vision systems using face recognition cannot.
The challenge is that animated characters have {\em huge} diversity. Although there is typically some anthropomorphic element, their appearance can range from human-like faces, to cars, fish, blobs, skeletons, \ldots (and that is just Pixar movies).
This long-tailed distribution of appearances poses a significant challenge for recognition methods. In contrast, (human) face recognition, which is the workhorse for character recognition in movies, is a far more constrained problem: 
a face has eyes, a nose, a mouth, maybe some
facial decorations (beard, glasses), in a regular configuration.  

In this paper, 
we present a computer vision pipeline for animated movies with a key design to recognise and track \emph{animated characters},
despite the long-tailed distribution of character appearance and exaggerated actions. We achieve this by using {\em visual} recognition methods to detect, segment, track and encode the appearance of the characters, and also {\em audio} recognition methods, since the voice of the animated character remains mostly the same throughout the movie.
We leverage online movie forums like IMDb and Fandom as external knowledge banks, and build a novel audio-visual character bank for animated movies automatically, including the voice samples of each voice actors. Both the visual and audio cues are then utilised to recognise and track animated characters effectively.



Recognising animated characters enables multiple downstream applications, and we develop two of these in this paper: Audio Description generation and character-aware subtitling, as illustrated in Fig.~\ref{fig:teaser}. Both of these applications depend heavily on character recognition.


In movies and TV series, \textit{Audio Descriptions (ADs)} are narrations that convey essential visual information to complement the original soundtrack. They help ensure a continuous and coherent narrative flow, enabling {\em visually-impaired} audiences to follow the plot effectively. ADs prioritise the most visually salient and story-relevant content such as character dynamics and significant objects, while omitting redundant details like background figures.



\textit{Character-aware subtitling}, on the other hand, concentrates on helping {\em hearing-impaired} people follow dialogues in video material. This actually requires determining three things for each utterance: \textit{what is being said}, \textit{when it is said}, and \textit{who is saying it}. By explicitly attributing dialogue to specific characters, it enhances the clarity of conversation flow, particularly in scenes with overlapping speech or multiple speakers.


In this paper, we make contributions over four areas:

\textbf{(i)} We develop an automatic pipeline for constructing an audio-visual character bank from online sources, consisting of a character appearance bank and a character voice bank (Sec.\ \ref{character_bank_construction}).

\textbf{(ii)} We tackle character recognition by leveraging both visual and audio information. For accurate visual animated character recognition, we propose Track-Guided Region Proposals (TGRP) and adapt DINOv2 to animated content. Additionally, we introduce visual enhancements to improve audio-only speaker recognition (Sec.\ \ref{av_char_recog}).

\textbf{(iii)} We use our animated character recognition to solve two downstream tasks -- Audio Description generation and character-aware subtitling; where previous face-based methods fail due to the large domain gap with animated content (Sec.\ \ref{down_stream}).

\textbf{(iv)} We introduce a new dataset, CMD-AM, consisting of $75$ animated movies with unified annotations, including character bounding boxes, ground truth Audio Descriptions, and sentence-level diarisation labels (Sec.\ \ref{sec:cmdam}).

Overall, our pipeline achieves superior character recognition accuracy compared to conventional human face detection methods. This, in turn, enables the generation of high-quality Audio Descriptions and character-aware subtitles, significantly enhancing the accessibility of animated movies.

\section{Related Work}

\begin{figure*}
    \centering
    \includegraphics[width=0.96\linewidth]{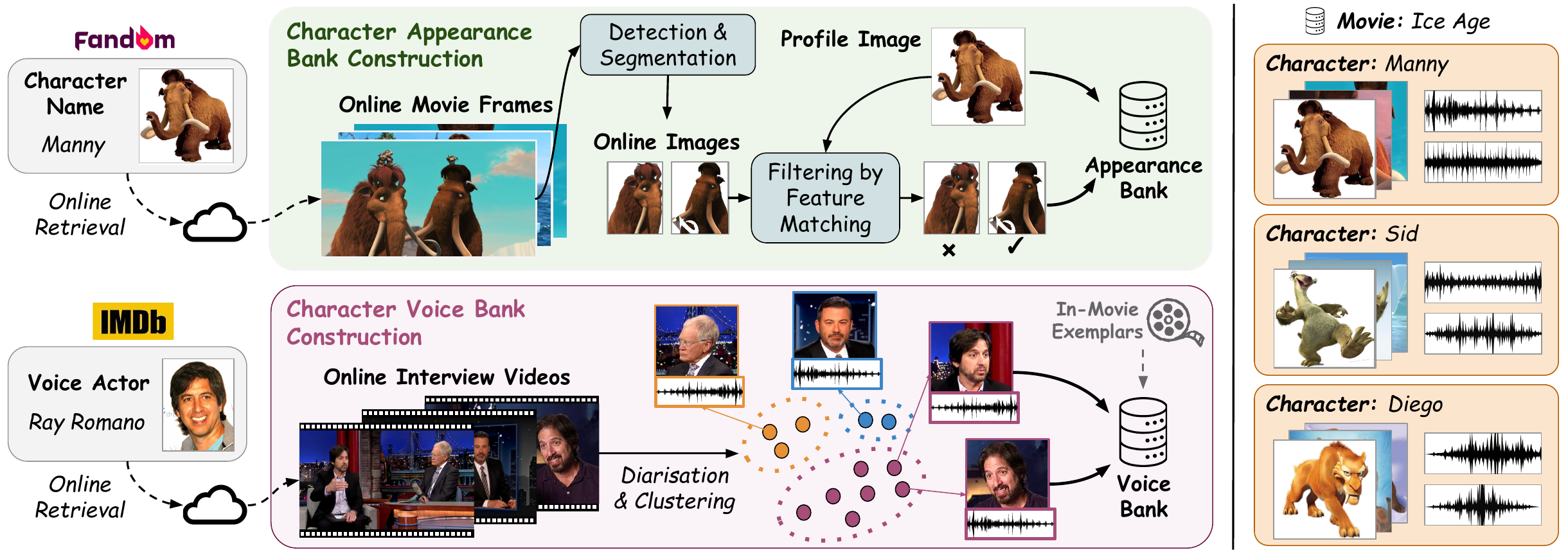}
    \caption{\textit{Audio-Visual Character Bank Construction.} \textit{Left:} 
    The audio-visual character bank is built automatically from online sources. Character names, corresponding voice actor names, and profile images are retrieved from IMDB and Fandom. We then construct a character appearance bank and character voice bank from online sources. \textit{Right:} An example audio-visual character bank for \textit{Ice Age}.}
    \label{fig:appearance_bank_construction}
\end{figure*}

\subsection{Animated Character Recognition}
Animated character recognition is often considered more challenging than real-world human identification, primarily due to large domain variations in animated media in terms of character geometry and styles. As a result, conventional human facial recognition methods~\cite{Deng_2019_CVPR, Kim_2022_CVPR, 10098610,Deng2020CVPR} often struggle with this task.

To address this issue, several datasets~\cite{9069265,zheng2020cartoon,danboorucharacter,rios2021dafre,Sachdeva24a} have been proposed for {\em image-based} animated character recognition, with follow-up research focusing on animated face detection~\cite{ogawa2018,8893046,zhang2020acfdasymmetriccartoonface} and recognition~\cite{Zheng2020,TIP.2022.3177952,10.1145/3321408.3322624,naftali2023aniwhoquickaccurate,10191980,10871249,soykan2023identity,10.1145/3638884.3638894}.

On the other hand, character recognition in animated {\em videos} remains relatively under-explored. Kim et al.~\cite{9133143} adapted Faster R-CNN~\cite{NIPS2015_14bfa6bb} for detecting characters in animated movies, while Somandepalli et al.~\cite{8017484} employed an unsupervised clustering approach to ``discover'' characters. More recently, CAST~\cite{cast2022} improved character clustering through self-supervised training of animation representations.
In contrast, our method automatically constructs character banks for target animated movies and formulates the recognition process as a few-shot retrieval task, thereby achieving scalable and generalisable animated character recognition.

\subsection{Audio Description Generation}
Audio Descriptions (ADs) are narrations in movies and TV series that provide visual information complementary to the audio, helping visually impaired individuals better understand the story. Existing AD generation approaches can be broadly divided into fine-tuned models~\cite{Han23,Han23a,Han24,movieseq,distinctad,uniad} and training-free methods~\cite{Zhang_2024_CVPR,chu2024llmadlargelanguagemodel,Xie24b,xie2025shotbyshot}. The former typically involve end-to-end architectures trained on ground truth AD sentences, while the latter adopts multi-stage strategies, leveraging pre-trained vision-language models (VLMs) and large language models (LLMs) as core components at different stages.

One of the key challenges in AD generation is accurately identifying the characters to be described. AutoAD-II~\cite{Han23a} first addressed this by training character prediction modules on existing movie datasets~\cite{MovieNet}. AutoAD-Zero~\cite{Xie24b} further improved accuracy by incorporating more robust off-the-shelf face recognition methods. However, these approaches are not directly applicable to animated movies due to the significant domain gap between real and animated characters. In this work, we propose a scalable and generalisable pipeline that accurately identifies animated characters and supports downstream AD generation.

\subsection{Character-Aware Subtitling}
Character-aware subtitling~\cite{Korbar24,Huh24b} is a recently emerging task that aims to generate subtitles for media such as movies and TV series while identifying the corresponding speakers. This significantly enhances the accessibility of video content for hearing-impaired people. The task generally involves two key steps:  
(i) Automatic speech recognition (ASR)~\cite{whisper,Bain23,radhakrishnan2023whispering,rouditchenko24_interspeech}, which transcribes spoken dialogue into text; and  
(ii) Speaker diarisation, which determines “who spoke when” in the audio stream.

In particular, speaker diarisation can be categorised into two main directions. The first focuses on audio-only diarisation~\cite{ICASSP.2018.8462628,zhang2019fully,kwon2021look,fujita2019end,horiguchi2020end,DiarizationLM}. The second direction, benefiting from recent advances in multimodal learning, leverages both audio and visual signals~\cite{mingote2024avdsurvey} to enhance diarisation performance. For instance, visual cues such as lip movements~\cite{he2022end,he2024qualityaware} and character faces~\cite{Chung20,10.1145/3503161.3548027,chung2019who,10.1145/3531232.3531250} have been shown to improve the accuracy of speaker attribution.
In this work, we address speaker diarisation for animated movies and demonstrate that accurate visual recognition of animated characters can significantly boost diarisation performance.






\section{Animated Character Bank Construction}
\label{character_bank_construction}
In animated movies, characters take on various forms; some resemble humans, while others have more abstract or imaginative appearances. Such a long-tailed distribution of character appearances presents a significant challenge to character recognition. 




In this section, we introduce an automatic pipeline for constructing an audio-visual character bank that captures both the appearance and corresponding voice of each character. 

As exemplified in Fig.\ \ref{fig:appearance_bank_construction} (right), for each movie, we maintain a character name list $\mathcal{S}=\{C^1,\ldots,C^N\}$, where the $p$-th character $C^p \in \mathcal{S}$ is associated with a tuple $\{I^{p}, A^{p}\}$,
where $I^{p}$ denotes a set of \(n\) exemplar images~\(I^{p} = \{I^{p}_1, \ldots, I_n^{p}\}\), and \(A^{p} = \{A^{p}_{1}, \ldots, A^{p}_{m}\}\) represents a set of \(m\) exemplar voice (audio) segments. In Sec.~\ref{appearance_bank}, we elaborate on the construction of the appearance bank, and in Sec.~\ref{voice_bank}, we detail the voice counterpart.

\subsection{Character Appearance Bank Construction}
\label{appearance_bank}

To construct the appearance bank for each movie, we first retrieve a character name list along with their profile images from online movie databases \cite{IMDb, Fandom}, as illustrated in Fig.~\ref{fig:appearance_bank_construction} (left). However, the set of profile images is often limited in size and typically depicts characters in a single, static pose, whereas animated characters exhibit a wide range of visual morphisms. To address this, we augment the character bank with additional images collected via web searching.


\paragraph{Candidate Image Retrieval.} 
Unlike profile images, web-crawled images often contain complex backgrounds and multiple characters. To isolate the character of interest, we extract all regions likely to contain relevant figures. Using an open-vocabulary detection model (OWLv2~\cite{owlv2}) to generate candidate boxes, we use these boxes as visual prompts for SAM2~\cite{ravi2024sam2} to segment potential character regions in the web-crawled images.

\paragraph{Image Filtering.} We then apply an additional filtering step to retain only the images corresponding to the target character. Specifically, for each character, we compute the cosine similarity between the DINOv2~\cite{oquab2024dinov2} feature of the profile image and each candidate character image. If the cosine similarity exceeds a predefined threshold, the candidate image is included in the appearance bank.


\paragraph{DINOv2 Adaptation.}
Nevertheless, we observe that pre-trained DINOv2 features often struggle to distinguish between visually similar characters, such as the Mammoths (Manny and Ellie) from \textit{Ice Age} in Fig. \ref{fig:appearance_bank_construction}.

To mitigate this, we further adapt the DINOv2 feature extractor on the character bank to deal with the long-tailed distribution of the appearance of animated characters. We individually fine-tune DINOv2 for each movie using our previously crawled images.


The fine-tuning process follows a contrastive training scheme. For each character \(C^p \in \mathcal{S}\), we randomly sample a positive pair \((I^p_i, I^p_j)\) from their exemplar images, while negative samples \(\{I^q_j\}\) are drawn from the remaining characters. Finally, we apply the standard InfoNCE~\cite{oord2018representation} loss to fine-tune the DINOv2 model. Full details are given in the Appendix.

\subsection{Character Voice Bank Construction}
\label{voice_bank}
In addition to the appearance bank, we also construct a voice bank containing multiple speech segments for each character. To obtain these voice exemplars, we consider two major sources: (i) online interview videos, and (ii) in-movie speech segments.

\paragraph{Interview Video Exemplar Extraction.} 
As shown in Fig.\ \ref{fig:appearance_bank_construction} (left), we first associate each character with their corresponding voice actor using information available from online movie databases. We then query search engines to retrieve multiple interview videos for each voice actor. To extract the speech segments of the target actor from these videos, we leverage the prior that the actor’s speech (as the interviewee) is likely to be the most frequently occurring voice across the interview videos. 

Specifically, we employ a speaker diarisation model~\cite{Bredin23} to partition the audio streams into multiple speaker clusters. We then select the largest resulting cluster and add its speech segments to the character voice bank.




\paragraph{In-Movie Exemplar Extraction.} 
However, compared to live-action movies, animated movies often exhibit a greater disparity between characters’ voices and the natural voices of their voice actors. To further mitigate this gap, we augment the character voice bank with in-movie speech segments, guided by our visual recognition results (detailed in Sec.~\ref{visual_recog}).

To be more specific, for each character, we select visually predicted tracks with the visual confidence scores higher than a defined threshold for the corresponding label. We then use an audio-visual synchronisation model~\cite{Afouras20b} to generate a synchronisation score to assess how likely the character on the track aligns with the concurrent speech. If the synchronisation score exceeds a predefined threshold, the corresponding in-movie speech segments are added to the character voice bank. For more details on the audio-visual synchronisation model, please refer to the Appendix.

\section{Audio-Visual Character Recognition}
\label{av_char_recog}
In this section, we address animated character recognition from two complementary perspectives: visual identification, detailed in Sec.~\ref{visual_recog}, and speaker attribution, described in Sec.~\ref{audio_recog}.


\subsection{Visual Recognition of Animated Characters}
\label{visual_recog}
Given the constructed character appearance bank (as detailed in Sec.\ \ref{appearance_bank}), we aim to visually recognise the animated character following a two-stage pipeline: (i) region proposal, which localises and tracks candidate character regions, and (ii) character identification, which assigns character IDs to each proposed track.


\begin{figure}[t]
    \centering
    \includegraphics[width=\linewidth]{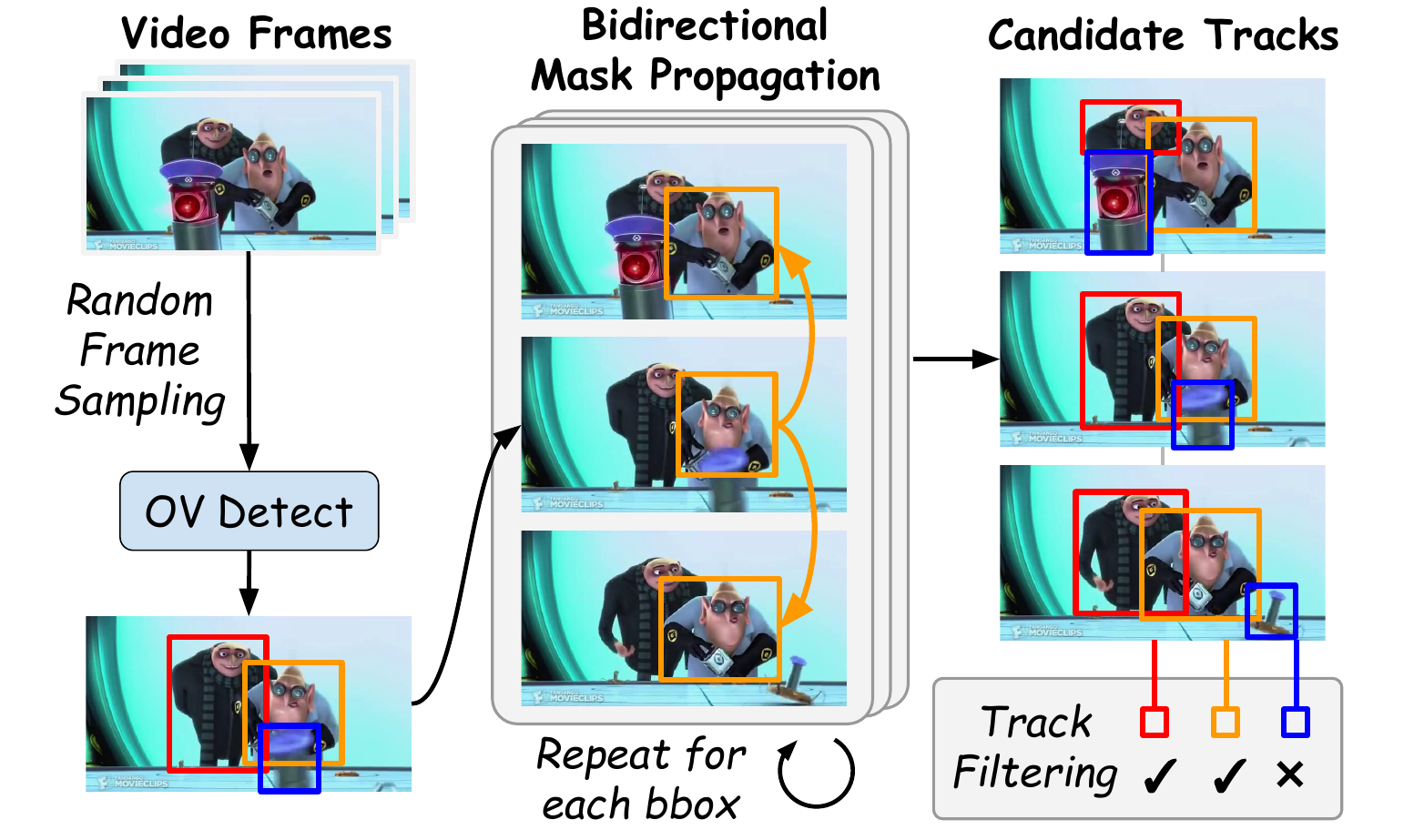}
    \caption{\textit{Track-Guided Region Proposals (TGRP).} We adopt an open-vocabulary detection (OV Detect) model to generate bounding box (bbox) proposals for sampled movie frames. These proposals are then propagated bidirectionally to produce candidate tracks, which are subsequently refined through a track filtering process.}
    \label{fig:region_proposals}
\end{figure}

\paragraph{Track-Guided Region Proposal.}
To ensure consistency across video clips, we propose a tracking-based approach for region proposal. As a preprocessing step, we apply a shot detection model to partition the video into individual shots, which naturally defines the boundaries for subsequent tracking operations.

As illustrated in Fig.~\ref{fig:region_proposals}, we randomly sample one seed frame from each shot and apply OWLv2~\cite{owlv2}, an open-vocabulary detection model, to generate coarse bounding box proposals. These proposals are then used to prompt SAM2~\cite{ravi2024sam2}, a segmentation and tracking model, which initialises a bi-directional mask propagation process to generate a set of candidate tracks.

We sample \(n = 3\) seed frames to repeat the process and generate a set of tracks for each seed. Then, for the track sets, we perform tripartite matching based on overlapping IoUs, considering a match valid if the IoU exceeds a predefined threshold. We discard any unmatched tracks, as they are likely to be outliers.

\paragraph{Character Identification.}
We use the adapted DINOv2 model (detailed in Sec.\ \ref{appearance_bank}) to assign character IDs to predicted tracks. Specifically, for each track, we uniformly sample \(5\) frames and compute their features \(\{f_j:j=1,\ldots,5\}\). To determine if the track matches with a character $C^p$, we calculate the cosine similarity between each query feature \(f_j\) and every exemplar feature \(z_i^p\) from the visual character bank. The visual matching score \(s_\text{vm}^p\) is then obtained by averaging over the top-\(k\) (\(k=3\)) cosine similarities, i.e.\ \(s_\text{vm}^p = \max_j\left(\frac{1}{k} \sum_{\text{top-}k,i} \langle f_j , z_i^p \rangle\right)\). Finally, we assign the track to the character with the highest matching score.

\subsection{Audio Recognition of Animated Characters}
\label{audio_recog}

Beyond visual recognition, we also aim to identify speakers in the audio stream, leveraging the pre-constructed character voice bank (detailed in Sec.\ \ref{voice_bank}). We start with an audio-only approach, and subsequently explore the incorporation of visual cues to further enhance speaker recognition.

\paragraph{Audio-Only Speaker Recognition.}
We begin with a straightforward audio-only speaker recognition approach, which matches speech segments in animated movies to voice exemplars in the character bank, thereby predicting speaker labels for each segment.

Given the audio stream of an animated movie, we first extract all speech segments using WhisperX~\cite{Bain23}. We then follow the strategy described in Sec.~\ref{voice_bank}, where speech segments are clustered using a diarisation model.
For each cluster, we extract its ``centroid feature'' by averaging ECAPA-TDNN~\cite{desplanques2020ecapa} voice embeddings of all contained speech segments.

To associate these clusters with characters in the voice bank, we adopt a feature-based matching process, similar to that introduced in Sec.~\ref{visual_recog}. Specifically, we use the cluster's centroid feature to query the ECAPA-TDNN features of different exemplar speech segments and compute their pairwise cosine similarities. For each character, the top-\(k\) (\(k=3\)) similarities are then recorded and averaged to obtain an audio matching score. We assign the cluster to the character with the highest score, \({s_{\text{am}}}\). Furthermore, for each speech segment in the cluster, we estimate a confidence score \(c_a\) by scaling \({s_{\text{am}}}\) with the cosine similarity between its feature and the centroid feature of the cluster.

\paragraph{Visual-Enhanced Speaker Recognition.}  
The proposed audio-only recognition, however, is prone to background noise commonly present in movies, such as music and environmental sounds. In addition, the character voice bank may also include noisy exemplars. To address these issues, we incorporate visual character recognition to enhance speaker identification.

To further enhance speaker recognition using visual cues, we focus on speech segments with confidence scores \(\{c_a\}\) below a predefined threshold, categorising them as low-confidence predictions.

\begin{figure}
    \centering
    \includegraphics[width=0.475\textwidth]{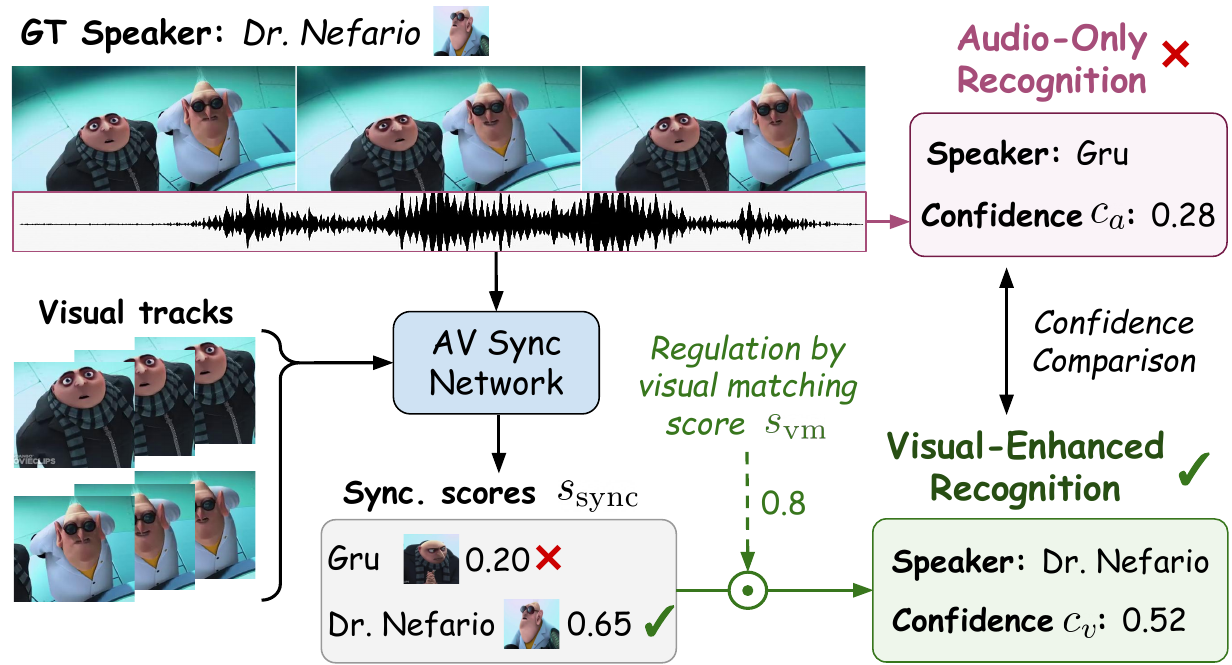}
    \caption{\textit{Visual-Enhanced Speaker Recognition.} We leverage an audio-visual synchronisation (AV Sync) network to compute synchronisation scores between simultaneous audio and character tracks. These scores are then used to determine whether a character is aligned with the audio signal, indicating that they are the speaker. The synchronisation scores are further regulated using visual matching scores. This results in a visual-enhanced speaker prediction and an associated visual confidence score. If the visual confidence exceeds the audio counterpart, we update the audio-only identified speaker with the visually recognised one.}
    \label{fig:av_correction}
\end{figure}

For each such audio segment, we retrieve all visually predicted character tracks that overlap with its time range, as illustrated in Fig.\ \ref{fig:av_correction}. 
For each retrieved track, we apply the same audio-visual synchronisation model we use to retrieve in-movie voice exemplars in Sec. \ref{voice_bank} to compute the synchronisation score \(s_\text{sync}\). This score indicates the likelihood that the track corresponds to the voice, specifically measuring the correlation between the temporal appearance variations of the tracked character and the temporal audio variations.

In addition, we incorporate the visual matching score \(s_\text{vm}\) (introduced in Sec.~\ref{visual_recog}) as a complementary measure of visual track reliability. We then combine the synchronisation and visual matching scores into a visual confidence score \(c_v = s_\text{sync} \cdot s_\text{vm}.\)
This score is scaled by a hyperparameter \(\lambda\) and compared to the audio-only confidence score \(c_a\). If \(\lambda \cdot c_v > c_a\), we interpret this as a more reliable visual-based prediction and update the audio-only prediction with the character ID inferred from the visual track.

\section{Applications for Animated Movies}
\label{down_stream}
Building on the visual and audio character recognition results, we explore downstream applications that enhance the understanding of animated movies. In particular, we focus on improving accessibility through two tasks: Audio Description (AD) generation (Sec.\ \ref{ad_gen}) and character-aware subtitling (Sec.\ \ref{char_subtitle}), targeting visually and hearing-impaired audiences, respectively.
\subsection{Audio Description Generation}
\label{ad_gen}
Audio Descriptions (ADs) are narrations that capture the most salient visual content in videos, serving as a complement to existing audio information. An AD generation method therefore takes video clips as input and aims to automatically generate AD sentences in textual form.

In this work, we build upon AutoAD-Zero~\cite{Xie24b}, a recent training-free AD generation framework. This method involves two major stages: In Stage I, a video-language model (VideoLLM) is prompted to generate a comprehensive description of the input video frames, covering key visual elements such as characters, actions, etc. In Stage II, a large language model (LLM) refines this dense text description into a concise and narratively coherent AD output.

Prior work~\cite{Han23a} has shown that accurate character recognition significantly enhances the quality of ADs. Accordingly, the original AutoAD-Zero~\cite{Xie24b} employs a robust face detection method (InsightFace~\cite{Deng_2019_CVPR}) for character identification. However, this approach underperforms in animated content, where conventional face recognition models struggle to detect animated characters.

To address this domain gap, we integrate our animated character bank and corresponding visual character recognition results into the AD generation framework. This is achieved by prompting the Stage I VideoLLM in a training-free manner. Specifically, we visually overlay the predicted character bounding boxes onto each input frame. Each character box is assigned a unique colour, and a text prompt provides the mapping between colours and character names. This allows the VideoLLM to ground the visual entities to their corresponding names and refer to them explicitly during the dense description generation.

\subsection{Character-Aware Subtitling}
\label{char_subtitle}

Character-aware subtitling aims to temporally localise spoken dialogues in movies, transcribe the corresponding audio, and assign each utterance to its corresponding speaker. This involves two main components: automatic speech recognition (ASR) and speaker diarisation.

\paragraph{Automatic Speech Recognition.}  
We utilise WhisperX to transcribe the video, producing individual dialogue segments along with their precise timestamps.

\paragraph{Speaker Diarisation.}  
Based on the transcribed timestamps, we extract the corresponding speech segments from the audio stream. These segments are then assigned to characters in the voice bank using the speaker recognition procedure described in Sec.\ \ref{audio_recog}.

Finally, each speaker label is paired with its corresponding dialogue transcription to form character-aware subtitles.

\section{The CMD-AM Dataset}
\label{sec:cmdam}
\begin{table}[t]
    \centering
    \caption{\textit{Overview of the CMD-AM dataset}, where ``AM'' stands for ``Animated Movie''. The dataset consists of 75 animated movies, with (partial) annotations including Audio Descriptions (ADs), per-frame character bounding boxes, and speaker labels for each audio segment.}
    \setlength\tabcolsep{3pt}
    \resizebox{0.475\textwidth}{!}{
    \begin{tabular}{lccl}
    \toprule
    Annotations & No. Movies & Duration ($h$) & No. Annotations \\ \midrule
    Audio Descriptions & $75$ & $24.5$ & $9,348$ ADs \\
    Character Boxes  & $8$ & $1.5$ & $126,977$ frames  \\
    Diarisation & $8$ & $1.5$ & $1,336$ audio segments \\
    \bottomrule
    \end{tabular}}
    \label{tab:dataset}
\end{table}

In this section, we introduce CMD-AM (``AM'' stands for ``Animated Movies''), a curated dataset comprising animated video clips sourced from the Condensed Movie Dataset (CMD)~\cite{Bain20}. In total, the dataset contains $565$ video clips from $75$ animated movies. The statistics are summarised in Tab.\ \ref{tab:dataset}.
CMD-AM provides a wide range of annotations including ground truth Audio Descriptions (Sec.\ \ref{anno_ad}), grounded bounding boxes for character recognition (Sec.~\ref{anno_box}), and sentence-level speaker labels for diarisation (Sec.~\ref{anno_speaker}). Note that these annotations are used \textit{for evaluation purposes only}, and our method is not trained on them.
The complete list of movies included in the CMD-AM dataset is available in the arXiv version.

\subsection{Ground Truth Audio Description}
\label{anno_ad}
We provide ground truth (GT) Audio Descriptions (ADs) for $565$ video clips in $75$ animated movies, covering a duration of $24.47$ hours. To obtain the GT AD annotations, we first collect AD-blended audio streams narrated by volunteers from AudioVault \cite{AudioVault}. We then follow the processing pipeline introduced in~\cite{Han24, Xie24b}, which extracts clean, text-form ADs from the downloaded audio streams through a series of transcription, diarisation, and post-filtering steps.

\subsection{Frame-Level Character Box Annotation}
\label{anno_box}
We provide frame-level bounding box annotations on characters for $40$ video clips across $8$ animated movies (listed in the arXiv version). In each clip, the video is sampled and labelled at $23.98$ fps, resulting in a total of $126,977$ annotated frames, covering approximately $1.5$ hours of video footage.

To obtain these frame-level bounding box annotations, we employ a three-stage process: (i) We first partition each video clip into multiple shots using an off-the-shelf shot detection model \cite{SceneDetect}. Within each shot, we randomly sample a few frames and manually annotate them with precise bounding boxes and corresponding character labels; (ii) We accelerate the annotation process by leveraging SAM2 to bidirectionally propagate the annotated boxes across each shot; (iii) Finally, we manually inspect every frame to correct any missing or erroneous boxes and labels, ensuring the overall accuracy of the annotations.

\subsection{Sentence-Level Diarisation Annotation}
\label{anno_speaker}
We provide sentence-level speaker labels for the \textit{same} $40$ video clips across $8$ movies, covering approximately $1.3$k speech segments.

During the curation process, we begin by manually identifying the speech segments for each character. These segments are then further divided at the sentence level, i.e.\ to correspond to individual sentences. Specifically, we leverage WhisperX to detect sentence boundaries and localise their corresponding timestamps. The segments are then split at these boundaries. Finally, we manually review and correct any timestamp errors to obtain high-quality sentence-level segments with corresponding speaker labels.

\section{Experiments}
In this section, we first provide implementation details of our method in Sec.\ \ref{exp:imple_detail}, followed by a comprehensive evaluation of its effectiveness. We assess performance on both visual and audio character recognition (i.e.\ character diarisation) in Sec.\ \ref{exp: visual_recog} and Sec.\ \ref{exp: audio_recog}, respectively. Additionally, we evaluate the effectiveness of our approach on downstream tasks, including Audio Description generation (Sec.\ \ref{exp: ad_gen}) and character-aware subtitling (Sec.\ \ref{exp: char_subtitle}).

\subsection{Implementation Details}
\label{exp:imple_detail}
More details are given in the arXiv version of this paper.

\paragraph{Character Appearance Bank.}  
For each movie, we select the top-10 characters from the cast list and collect 8 exemplar images per character from online sources. Character regions are detected using OWLv2~\cite{owlv2} with the prompt \emph{``a photo of animated character''} and segmented via SAM 2.1~\cite{ravi2024sam2}. Manual correction is performed when web-crawled images are incorrect.

To adapt DINOv2 to animated content, we fine-tune DINOv2 ViT-g/14 (w/ 4 register tokens)~\cite{oquab2024dinov2} by updating its final linear layer. Training lasts for 75 epochs with a cosine annealing learning rate schedule from $6\times10^{-4}$ to $5\times10^{-6}$, using InfoNCE loss with temperature $\tau=0.07$.

\paragraph{Character Voice Bank.}  
To build voice exemplars, we collect 5 interview videos per voice actor. In-movie audio segments are selected when the visual match score $s_{\mathrm{vm}}>0.6$ and the sync score $s_{\mathrm{sync}}>0.3$, computed using an adapted version of LWTNet~\cite{Afouras20b}.
In our character voice bank, each character has 15 voice segments; if a character has fewer than 15 in-movie samples, we supplement them with segments derived from interview audio.

\paragraph{Visual Character Recognition.}  
We detect shot boundaries 
using SceneDetect~\cite{SceneDetect}, then generate region proposals with OWLv2 and track them using SAM 2.1. Tracks are matched if their IoU is at least $0.5$. Frame-level predictions are refined using non-maximum suppression (NMS) with IoU threshold $0.5$.

\paragraph{Audio Character Recognition.}  
Audio-only predictions with confidence $c_{a} < 0.35$ are refined using synchronised visual tracks. We compute a composite score (Sec.~\ref{audio_recog}) with $\lambda=1.0$ and update the predicted label to that of the highest-scoring concurrent track.

\paragraph{Audio Description Generation.}  
Following AutoAD-Zero, we sample 8 frames from the AD interval. Character predictions with $s_{\mathrm{vm}}>0.45$ are retained. Bounding boxes around the characters' bodies are overlaid to indicate character identity, using unique colours and names in the prompt.

\begin{table}[t]
    \centering
    \caption{\textit{Visual Character Recognition on CMD-AM.} ``DINOv2 adap.'' denotes DINOv2 adaptation. Note, *: 
    our bounding box annotations are for entire character bodies (as the facial boundaries of animated characters are often ambiguous), while InsightFace detects face boxes. This mismatch results in low IoUs between the two sources, leading to the reported 0.0 mAP value.
    }
    \setlength\tabcolsep{3pt}
    \resizebox{0.475\textwidth}{!}{
    \begin{tabular}{cccccc}
    \toprule
     \multirow{2}{*}{Exp.} & \multirow{2}{*}{Method}   & \multirow{2}{*}{\shortstack{DINOv2 \\ adap.}} & \multirow{2}{*}{Tracking} &  \multirow{2}{*}{\shortstack{Char. Name \\  AP $\uparrow$}} & \multirow{2}{*}{\shortstack{Char. Bbox \\ mAP $\uparrow$}} \\
     & & & &  &  \\ 
     \midrule
     A & InsightFace~\cite{Deng_2019_CVPR} & $-$ & $-$  & $45.3$ & $0.0$* \\
     B & Ours & \ding{55} & \ding{55} & $75.3$ & $30.9$ \\
     C & Ours & \ding{51} & \ding{55} & $\mathbf{81.8}$ & $35.2$ \\
     \midrule
     \textbf{D} & Ours & \ding{51} & \ding{51} & $80.1$ & $\mathbf{45.3}$\\
     \bottomrule
    \end{tabular}}
    \label{tab:visual_recog}
\end{table}

\subsection{Visual Animated Character Recognition}
\label{exp: visual_recog}
We evaluate visual animated character recognition on the CMD-AM dataset from two perspectives: (i) whether the character names in a given clip are correctly predicted, and (ii) whether the character positions are accurately detected. The former is measured using Average Precision (AP) between the predicted and ground truth name lists for $3$-shot clips. For character detection, we evaluate the predicted bounding boxes and report the mean Average Precision (mAP), calculated by averaging AP scores at IoU thresholds ranging from $0.50$ to $0.95$ in increments of $0.05$. We use full-body bounding boxes for characters instead of face boxes, because animated characters often have ambiguous facial boundaries, as seen with Lightning McQueen in \textit{Cars}.



We report the quantitative results in Tab.\ \ref{tab:visual_recog}, where the baseline InsightFace method is found to struggle with recognising animated characters. In contrast, our visual recognition approach yields significant improvements. In particular, we observe a notable performance boost when adapting DINOv2 on the animated character bank, as seen by comparing Exp.\ B and Exp.\ C. Furthermore, incorporating the Track-Guided Region Proposal (TGRP) scheme (Exp.\ D) leads to comparable character name identification results with per-frame detection (Exp.\ C), while achieving significantly higher accuracy in grounded bounding box predictions.

\begin{table}[t]
    \centering
    \caption{\textit{Audio Character Recognition on CMD-AM.} We report Average Precision (AP) for various methods while predicting and evaluating speaker labels using ground truth timestamps.}
    \resizebox{0.4\textwidth}{!}{
    \begin{tabular}{cccc}
    \toprule
     In-Movie Exemplar & Visual Enhanced & Speaker AP $\uparrow$ \\ \midrule
     \ding{55} & \ding{55} & $73.8$ \\
     \ding{51} & \ding{55} & $77.3$ \\
     \midrule
     \ding{51} & \ding{51} & $\mathbf{77.9}$ \\
     \bottomrule
    \end{tabular}
    }
    \label{tab:audio_recog}
\end{table}

\subsection{Audio Animated Character Recognition}
\label{exp: audio_recog}
We evaluate our method's performance on audio-only animated character recognition. The evaluation is conducted on $8$ animated movies from CMD-AM, each annotated with sentence-level speaker diarisation labels. We report the Average Precision (AP) between the predicted and ground truth speaker labels at the sentence level.


\begin{table*}[t]
    \centering
    \setlength\tabcolsep{6pt}
    \caption{\textit{Audio Description Generation on CMD-AM.} ``TGRP'' is short for Track-Guided Region Proposal, and ``DINOv2 adap.'' denotes DINOv2 adaptation. The LLM-AD-Eval scores are evaluated using LLaMA2-7B (left) and LLaMA3-8B (right).}
    \resizebox{0.975\textwidth}{!}{
    \begin{tabular}{cccccccc}
    \toprule
     Method & Appearance Bank & Character Recognition & VLM & LLM & CRITIC\,$\uparrow$ & CIDEr\,$\uparrow$ & LLM-AD-Eval\,$\uparrow$ \\ \midrule
     AutoAD-Zero & \ding{55} & No Label & VideoLLaMA2-7B~\cite{damonlpsg2024videollama2} & LLaMA3-8B~\cite{grattafiori2024llama3herdmodels} & $2.3$ & $9.4$ & $2.64|\mathbf{1.46}$\\
     AutoAD-Zero & \ding{51} & InsightFace & VideoLLaMA2-7B~\cite{damonlpsg2024videollama2} & LLaMA3-8B~\cite{grattafiori2024llama3herdmodels} & $15.0$ & $11.3$ & $\mathbf{2.69}|\mathbf{1.46}$\\
     Ours & \ding{51} & TGRP + DINOv2 adap. & VideoLLaMA2-7B~\cite{damonlpsg2024videollama2} & LLaMA3-8B~\cite{grattafiori2024llama3herdmodels} &$\mathbf{28.6}$ & $\mathbf{12.2}$ & $2.68|1.43$\\ \midrule
     AutoAD-Zero & \ding{55} & No Label & Qwen2-VL-7B~\cite{Qwen2VL} & LLaMA3-8B~\cite{grattafiori2024llama3herdmodels} & $10.0$ & $13.0$ & $2.91|\mathbf{1.84}$ \\
     AutoAD-Zero & \ding{51} & InsightFace & Qwen2-VL-7B~\cite{Qwen2VL} & LLaMA3-8B~\cite{grattafiori2024llama3herdmodels} & $18.9$ & $14.3$ & $2.94|1.83$ \\
     Ours & \ding{51} & TGRP +  DINOv2 adap.  & Qwen2-VL-7B~\cite{Qwen2VL} & LLaMA3-8B~\cite{grattafiori2024llama3herdmodels} &$\mathbf{30.1}$ & $\mathbf{15.5}$ & $\mathbf{2.96|1.84}$ \\
     \bottomrule
    \end{tabular}}
    \label{tab:ad_gen}
\end{table*}

\begin{figure*}[ht]
    \centering
    \includegraphics[width=0.975\linewidth]{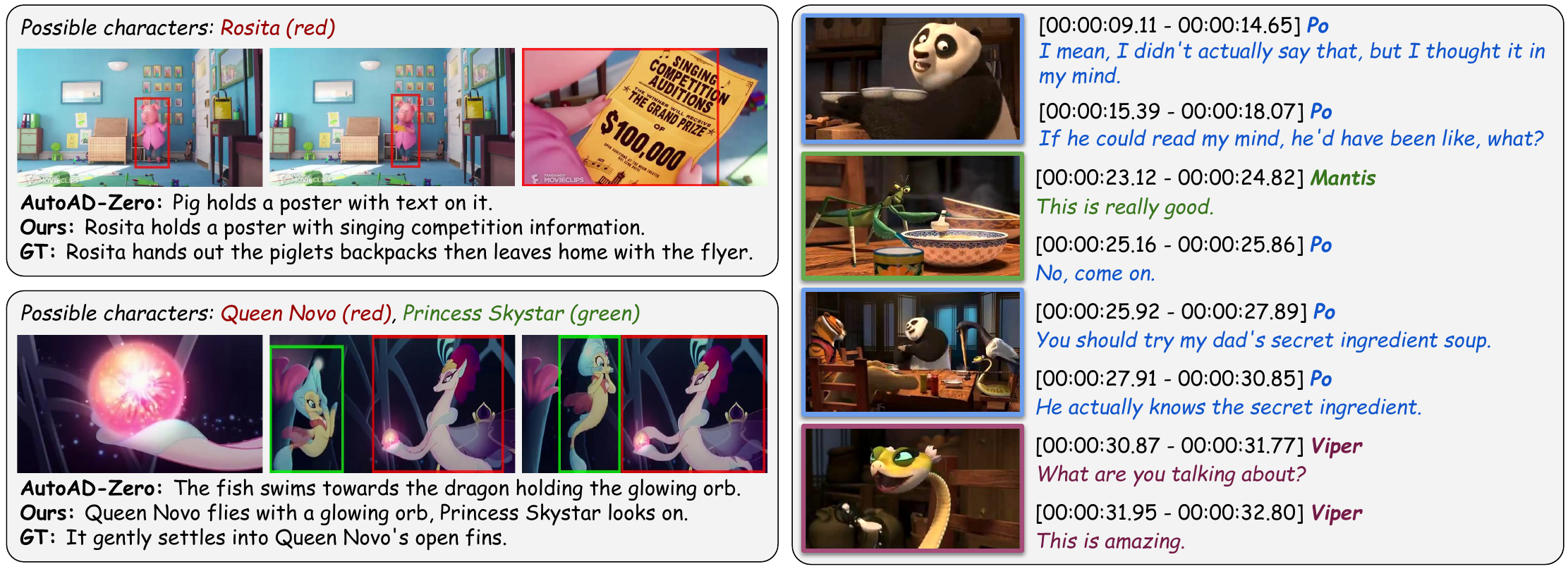}
    \caption{\textit{Qualitative Visualisations on CMD-AM.} 
    \textit{Left:} Predicted Audio Descriptions, where recognised characters are highlighted with coloured bounding boxes, and their names along with corresponding colour codes are provided in the text prompts. The default VLM used is Qwen2-VL-7B. Examples are taken from \textit{Sing} (top) and \textit{My Little Pony: The Movie} (bottom). \textit{Right:} Predicted character-aware subtitles for a segment from \textit{Kung Fu Panda}.}
    \label{fig:visualization}
\end{figure*}


As shown in Tab.\ \ref{tab:audio_recog}, augmenting the character voice bank with in-movie exemplars improves recognition accuracy compared to using voice exemplars sourced solely from interview videos. 
Furthermore, visual-enhanced recognition corrects errors made by the audio-only approach, leading to more accurate speaker identification.


\subsection{Audio Description Generation}
\label{exp: ad_gen}
To assess the quality of generated ADs, we follow previous works~\cite{Han24} and report the performance on CRITIC, CIDEr, and LLM-AD-Eval. Specifically, CRITIC~\cite{Han24} measures the accuracy of character identification in predicted ADs. CIDEr~\cite{Vedantam_2015_CVPR} evaluates text description quality based on the TF-IDF~\cite{Robertson2004}-weighted word matching between predicted and GT ADs. LLM-AD-Eval~\cite{Han24} leverages large language models (LLaMA2-7B~\cite{touvron2023llama2openfoundation} $|$ LLaMA3-8B~\cite{grattafiori2024llama3herdmodels}) to assess the quality of generated ADs, scoring them from 1 (lowest) to 5 (highest).

We compare our approach against AutoAD-Zero under two configurations: a baseline version without any character label information, and a variant that uses InsightFace combined with our constructed character bank for character recognition. We exclude other training-dependent methods, particularly those explicitly tuned on the CMD-AD~\cite{Han24} training set, since our CMD-AM dataset partially overlaps with it. We evaluate our method on all $75$ animated movies in the CMD-AM dataset.

We report the performance of the AD generation framework leveraging two different base VLM models, namely VideoLLaMA2-7B~\cite{damonlpsg2024videollama2} and Qwen2-VL-7B~\cite{Qwen2VL}. As shown in Tab.\ \ref{tab:ad_gen}, incorporating our character appearance bank and tracking-based character recognition method improves the quality of predicted ADs, especially CRITIC scores, indicating more accurate character references in the generated descriptions.


Fig.\ \ref{fig:visualization} (left) presents two examples of our predicted ADs. With our method, each AD correctly identifies the character names and exhibits improved description quality compared to AutoAD-Zero. More visualisations are given in the Appendix.


\begin{table}[t]
    \centering
    \caption{\textit{Character-Aware Subtitling on CMD-AM.} Diarisation Error Rates (DER) (in \%) are reported. ``DER (O)'' indicates DER including overlapping speech, whereas ``DER'' excludes overlapping speech.}
    \begin{tabular}{ccc}
    \toprule
        Visual Enhanced & DER (O) $\downarrow$ & DER $\downarrow$ \\
    \midrule
        \ding{55} & 40.3 & 40.0 \\
        \ding{51} & \textbf{35.8} & \textbf{35.7}\\
    \bottomrule
    \end{tabular}
    \label{tab:subtitling}
\end{table}

\subsection{Character-Aware Subtitling}
\label{exp: char_subtitle}
For character-aware subtitling, we measure speech recognition performance using Diarisation Error Rate (DER), which is calculated as the sum of missed speech, false alarms, and speaker confusion errors, divided by the total reference speech time. We report DER both with (DER(O)) and without (DER) overlapping speech. Moreover, we evaluate closed-set performance by excluding segments with ground truth speakers not present in the character bank, as well as segments corresponding to characters singing in \textit{Sing}.

Tab.\ \ref{tab:subtitling} presents our character-aware subtitling results. By incorporating visual-enhanced speaker recognition, the error rates get further reduced. Fig.\ \ref{fig:visualization} (Right) illustrates character-aware subtitles for a segment from \textit{Kung Fu Panda}. Our method produces accurate timestamps and correct character labels, resulting in high-quality subtitles. Further examples are given in the Appendix.

\section{Discussion}
Computer vision systems have been developed over decades for human facial recognition, 
and are trained on millions of identities/images,
while the challenging task of recognising free-form animated characters remains under-explored. Our method achieves accurate audio-visual character recognition with the aid of automatically constructed character banks, enabling character-centric understanding of animated movies for applications such as Audio Description generation and character-aware subtitling.


A limitation of our method is when there are multiple identical animated characters in the same shot. For instance, when there are numerous Minions presenting simultaneously, distinguishing between individual instances becomes challenging.
Other than that, the current limitations of our pipeline primarily lie in its downstream tasks. For instance, AutoAD-Zero does not account for temporal context, making it difficult to incorporate story-level plot elements into AD generation. Additionally, when it comes to character-aware subtitling, existing methods often struggle to accurately associate ``special'' speech segments with the correct speakers, such as those containing singing, crying, or screaming -- which are particularly common in animated videos.

\section*{Acknowledgements}
This research is supported by UK EPSRC Programme Grant VisualAI (EP/T028572/1), a Royal Society Research Professorship RSRP$\backslash$R$\backslash$241003, and a Clarendon Scholarship.
\bibliographystyle{ACM-Reference-Format}
\balance
\bibliography{reference, vgg_local}

\clearpage
\appendix
\twocolumn[
\begin{center}
\Huge \textbf{Character-Centric Understanding of Animated Movies} \\ \vspace{0.2cm}
\huge Appendix
\end{center}
\vspace{0.4cm}
]
\setcounter{page}{1}

\renewcommand{\thefigure}{A\arabic{figure}} 
\setcounter{figure}{0} 
\renewcommand{\thetable}{A\arabic{table}}
\setcounter{table}{0} 

This Appendix contains the following sections:
\begin{itemize}
    \item In Sec.\ \ref{supsec:add_imple}, we elaborate on \textbf{additional implementation details}, including the construction of the character voice bank and adaptations of pre-trained models to animated movies.
    \item In Sec.\ \ref{supsec:ablation}, we conduct \textbf{ablation studies} to verify key design choices and hyperparameter settings.
    \item In Sec.\ \ref{supsec:hardware}, we specify the hardware requirements and inference times.
    \item In Sec.\ \ref{supsec:cmdad_vis}, we present \textbf{more details on CMD-AM}, including the specific movie list and the visualisations on the ground truth character bounding boxes.
    \item In Sec.\ \ref{supsec:qualitative}, we provide \textbf{additional qualitative results} for our predicted ADs and character-aware subtitles.
\end{itemize}



\section{Additional Implementation Details}
\label{supsec:add_imple}
In this section, we provide additional implementation details of our pipeline. In Sec.\ \ref{supsubsec:interview_exemplar}, we provide a detailed explanation of the clustering process used for extracting interview video exemplars during character voice bank construction. In Sec.\ \ref{supsubsec:av_sync}, we elaborate on the adaptation of the audio-visual synchronisation network for animated movies. Finally, in Sec.\ \ref{supsubsec:dinov2_ttf}, we present a mathematical formulation of the DINOv2 adaptation process.

\subsection{Interview Video Exemplar Extraction}
\label{supsubsec:interview_exemplar}
For each character, we crawl $5$ interview videos, denoted as \(V = \{v_1,\ldots, v_5\}\). For each interview video \(v_i \in V\), we use a diarisation model to segment the audio stream into distinct speaker clusters based on the model-assigned labels, denoted as \(C_i = \{c_{i1}, \ldots, c_{ij}\}\). We then group the clusters from the crawled interview videos to form the set \(C = \{c_{ij} \mid i \in I, j \in J\}\). For each cluster, we compute its centroid feature, resulting in the set \(Z = \{z_{ij} \mid i \in I, j \in J\}\).

Based on the prior that the actor’s speech (as the interviewee) is likely to be the most frequently occurring voice across the interview videos, we implement a coarse merging of the speaker clusters, as shown in Alg. \ref{alg:merging_clusters}, and add the speech segments in the largest resulting cluster into the character voice bank.


\begin{algorithm}
\caption{Merging Speaker Clusters}
\label{alg:merging_clusters}
\begin{algorithmic}[1]
\STATE \textbf{Input:} \( Z = \{z_{ij} \mid i \in I, j \in J\} \), cosine similarity threshold \(\tau\)
\STATE \textbf{Output:} Merged Clusters \( \mathcal{M} \)
\STATE Initialize clusters \( \mathcal{M} \gets \emptyset \)
\FOR{$z \in Z$}
    \STATE $assigned \gets \textbf{false}$
    \FOR{$m \in \mathcal{M}$}
        \STATE $max\_similarity \gets \max_{c \in m}\bigl(\text{cosine\_similarity}(z, c)\bigr)$
        \IF{$max\_similarity > \tau$}
            \STATE $m$.append($z$); \quad $assigned \gets \textbf{true}$; \quad \textbf{break}
        \ENDIF
    \ENDFOR
    \IF{\NOT $assigned$}
        \STATE $\mathcal{M}$.append($\{z\}$)
    \ENDIF
\ENDFOR
\STATE \textbf{return} \( \mathcal{M} \)
\end{algorithmic}
\end{algorithm}

\subsection{Adapting Audio-Visual Synchronisation Network for Animated Contents}
\label{supsubsec:av_sync}

\begin{figure}[t]
    \centering
    \includegraphics[width=\linewidth]{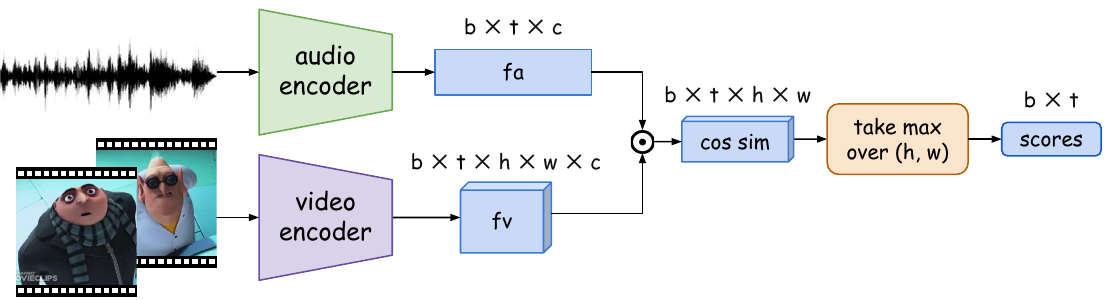}
    \caption{\textit{Generating Synchronisation Score.} We compute a synchronisation score between the on-screen character tracks and the concurrent audio stream to accurately detect the active speaker.}
    \label{fig:synchronisation_score}
\end{figure}

To establish audio-visual correspondence, we adopt LWTNet~\cite{Afouras20b}, which grounds character voices to their visual presence in each frame. As shown in Fig. \ref{fig:synchronisation_score}, the model consists of two branches which encode spatiotemporal visual features \(f_v \in \mathbb{R}^{t \times h \times w \times c}\) and audio features \(f_a \in \mathbb{R}^{t \times c}\), respectively. Cross-modal cosine similarities are computed across time, resulting in a similarity map of dimension \(t \times h \times w\). We then define a synchronisation score by taking the maximum over spatial dimensions and averaging across the temporal dimension.

\begin{figure}[t]
    \centering
    \includegraphics[width=\linewidth]{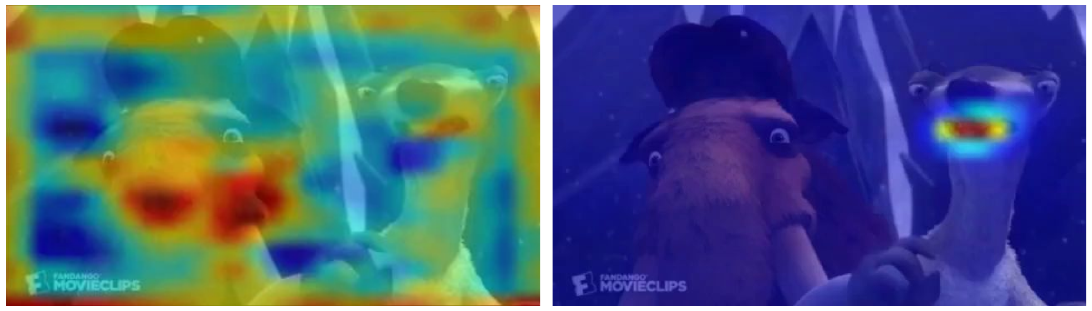}
    \caption{\textit{Audio-Visual Attention Visualisation.} \textit{Left:} Attention visualisation of pre-trained LWTNet on \textit{Ice Age}. \textit{Right:} Attention visualisation of fine-tuned LWTNet with longer temporal windows.}
    \label{fig:attn_vis}
\end{figure}

However, we found that the original LWTNet, pre-trained on real-world videos, generalises poorly to animated content. This limitation likely stems from the slower and less obvious lip movements of animated characters compared to real humans. Afouras et al. \cite{Afouras20b} fine-tune on large volumes of domain-specific animated content, such as episodes from \textit{The Simpsons}, to mitigate the domain gap. In contrast, our approach targets multiple animated movies with diverse character designs while contending with very limited video data for each movie. To address this, we fine-tune LWTNet with a larger temporal granularity on real videos, enabling it to perceive lip movements over longer temporal windows, and transfer directly to animated content.

To be more specific, the original LWTNet aggregates information from frames within $0.2$s, and we extend the temporal granularity to $0.6$s. This is achieved by adding a temporal adapter to video encoder and audio encoder, respectively. We use a convolution layer as our adapter to fuse the information within extended temporal windows.

For fine-tuning the audio-visual synchronisation model, we follow the training of LWTNet and consider synchronisation as a proxy task. We fine-tune the last two linear layers in both encoders in LWTNet and the temporal adapters on LRS2 and LRS3 datasets~\cite{Chung17}. Additionally, we generate synthetic videos by concatenating a static head with a talking head from the LRS datasets, addressing the scarcity of examples in the LRS datasets that feature multiple people in the frame. As illustrated in Fig. \ref{fig:attn_vis}, after fine-tuning with a larger temporal granularity, the audio-visual attention focuses on Sid's mouth -- the region associated with the active speaker.

\subsection{DINOv2 Adaptation}
\label{supsubsec:dinov2_ttf}
To adapt the DINOv2 feature extractor to animated content, we follow a contrastive training scheme. For each character \(C^p \in \mathcal{S}\), we randomly sample a positive pair \((I^p_i, I^p_j)\) from their exemplar images, while negative samples \(\{I^q_j\}\) are drawn from the remaining characters. We sample one image for each character \(C_q \in \mathcal{S} \setminus C_p\) to form the negative samples. Finally, we apply the standard InfoNCE~\cite{oord2018representation} loss to finetune the DINOv2 model. Formally, 
\begin{equation}
    \mathcal{L} = -\!\!\sum_{C^p \in \mathcal{S}}\!\log\frac{\exp \left((z^{p}_i\!\cdot\! z^{p}_j) /\tau\right)}{\exp\left((z^{p}_i\!\cdot\! z^{p}_j)/\tau \right) + \sum_{C^q \neq C^p}\exp\left((z^{p}_i\!\cdot\!z^{q}_j)/\tau\right)}
\label{infonce}
\end{equation}
where \(z_{i}=\text{DINOv2}(I_i)\) denotes the normalised CLS feature of the exemplar image \(I_i\) and \(\tau\) is the temperature parameter.
In our implementation, \(\tau\) is $0.07$.

\section{Ablation Study}
\label{supsec:ablation}
We conduct ablation studies on the manual curation of the character appearance bank for visual animated character recognition and on various design choices for audio animated character recognition. In particular, we investigate the impact of clustering and the number of audio exemplars included in the character bank.

\paragraph{Manual Curation of Character Appearance Bank.} 
Instead of relying on automatic web searches, we manually crawl online images for two movies in CMD-AM: \textit{The Polar Express} and \textit{Spider-Man: Into the Spider-Verse}. The remainder of the automatic character bank construction pipeline remains unchanged.

As shown in Tab.\ \ref{tab:ablation_charbank}, such manual curation of two problematic movies further improves the character recognition performance.


\begin{table}[t]
    \centering
    \caption{\textit{Ablation on Manual Curation of Character Appearance Bank.} Here, manual curation specifically refers to the manual retrieval (instead of automatic crawling) of online images as sources for constructing the character appearance bank.}
    \resizebox{0.475\textwidth}{!}{
    \begin{tabular}{cccc}
    \toprule
     Tracking & Manual Curation & AP@.5 $\uparrow$ & mAP@[.5,.95] $\uparrow$ \\ \midrule
     \ding{55} & \ding{55} & $39.9$ & $30.4$ \\
     \ding{55} & \ding{51} & $44.4$ & $35.2$ \\ \midrule
     \ding{51} & \ding{55} & $42.3$ & $37.2$ \\
     \ding{51} & \ding{51} & $52.2$ & $45.3$ \\
     \bottomrule
    \end{tabular}}
    \label{tab:ablation_charbank}
\end{table}

\begin{table}[t]
    \centering
    \caption{\textit{Ablation on Speaker Recognition for Animated Characters.} We compare speaker recognition APs across different numbers of voice exemplars in the character voice bank and assess the performance of a clustering-based classification versus direct feature matching approach.}
    \begin{tabular}{ccc}
    \toprule
     Clustering & Number of Exemplars $n$ & Speaker AP $\uparrow$  \\ \midrule
     \ding{55} & $10$ & $76.2$\\
     \ding{55} & $15$ & $76.4$\\
     \ding{55} & $30$ & $76.4$\\
     \midrule
     \ding{51} & $15$ & $77.3$\\
     \bottomrule
    \end{tabular}
    \label{tab:ablation_audio}
\end{table}

\paragraph{Speaker Recognition for Animated Characters.} In Tab. \ref{tab:ablation_audio}, we conduct an ablation study on our speaker recognition approach. We compare different numbers of exemplars, $n$, in the character voice bank, and assess the performance of a clustering-based classification method versus direct feature matching. Our results indicate that clustering allows us to more effectively exploit the voice similarity among adjacent instances of the same character. Augmenting the character voice bank with in-context exemplars also helps mitigate sample imbalance across characters in the original bank.

\section{Hardware Requirements}
\label{supsec:hardware}
All experiments can be conducted on a single NVIDIA A40/A6000 GPU (48GB), which supports both DINOv2 test-time fine-tuning and inference with 7B/8B VLMs and LLMs. Evaluating our method on the CMD-AM dataset takes approximately two days on eight NVIDIA A40 GPUs (48GB).

\section{More details on CMD-AM}
\label{supsec:cmdad_vis}
\begin{figure*}[t]
    \centering
    \includegraphics[width=0.92\linewidth]{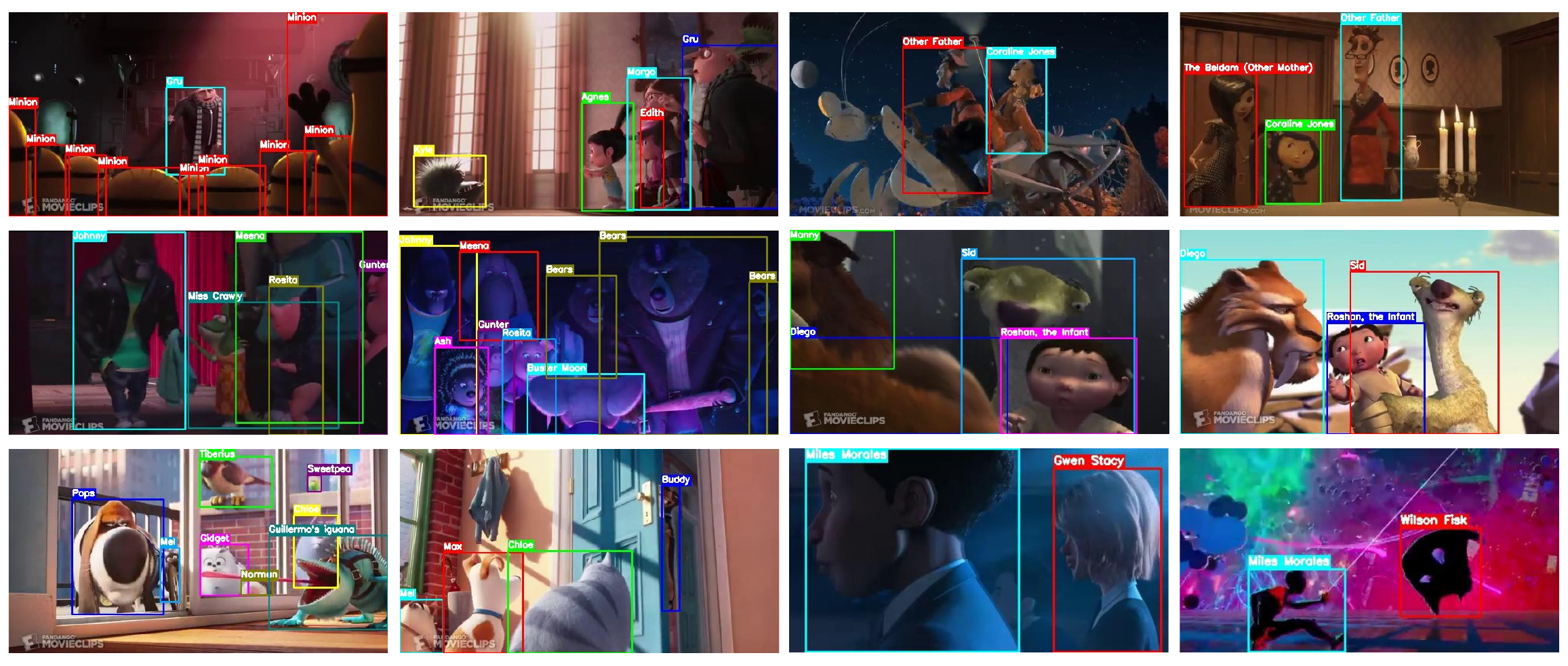}
    \caption{Visualisation of Our Character Box Annotations in CMD-AM.}
    \label{fig:cmdam}
\end{figure*}

We include a list of all animated movies in the CMD-AM dataset, as shown in Tab.\ \ref{tab:movie_names}.
We also provide exemplary ground truth annotations of character bounding boxes in Fig.\ \ref{fig:cmdam}.

\begin{table*}[t]
    \centering
    \caption{List of Animated Movies in the CMD-AM Dataset}
    \begin{tabular}{p{0.04\textwidth} p{0.05\textwidth} p{0.05\textwidth} p{0.775\textwidth}}
    \toprule
         GT AD & GT Bbox & GT Subt. & \multirow{2}{*}{Movie names} \\
         \midrule
         \ding{51} &\ding{51} & \ding{51} & \textbf{8 movies:} \textit{Coraline, Despicable Me, Ice Age, Kung Fu Panda, Spider-Man: Into the Spider-Verse, Sing, The Secret Life of Pets, The Polar Express} \\
         \midrule
         \ding{51} &\ding{55} & \ding{55} & \textbf{67 movies:} \textit{An American Tail, The Land Before Time, All Dogs Go to Heaven,
         An American Tail: Fievel Goes West, Chicken Run, The Prince of Egypt, Shrek, The Iron Giant, The Rugrats Movie, The Road to El Dorado, Joseph: King of Dreams, Ice Age, Jimmy Neutron: Boy Genius, Shrek 2, Looney Tunes: Back in Action, Coraline, The Polar Express, The SpongeBob SquarePants Movie, Astro Boy, Monster House, Bee Movie, Open Season, Shrek the Third, Barnyard, Surf's Up, Flushed Away, Kung Fu Panda, Puss in Boots, Horton Hears a Who!, The Croods, Planet 51, The Boxtrolls, Hotel Transylvania, Cloudy with a Chance of Meatballs, How to Train Your Dragon, Monsters vs. Aliens, Shrek Forever After, Megamind, Rango, Alpha and Omega, Legend of the Guardians, Kung Fu Panda 2, Despicable Me, Arthur Christmas, The Pirates! Band of Misfits, Rio, Rise of the Guardians, ParaNorman, How to Train Your Dragon 2, Trolls, Despicable Me 2, Sausage Party, Turbo, Penguins of Madagascar, Cloudy with a Chance of Meatballs 2, Captain Underpants: The First Epic Movie, Kung Fu Panda 3, The SpongeBob Movie: Sponge Out of Water, Minions, Sherlock Gnomes, How to Train Your Dragon 3, Smurfs: The Lost Village, Hotel Transylvania 2, The Grinch, The Secret Life of Pets, Despicable Me 3, Sing, The Boss Baby, My Little Pony: The Movie, Kubo and the Two Strings, Spider Man: Into the Spider-Verse, The Secret Life of Pets 2, The Angry Birds Movie 2, Abominable, Teen Titans GO! to the Movies}  \\ 
    \bottomrule
    \end{tabular}
    \label{tab:movie_names}
\end{table*}

\section{Additional Qualitative Results}
\label{supsec:qualitative}
\subsection{Qualitative Results of ADs}
\begin{figure*}[ht]
    \centering
    \includegraphics[width=0.95\linewidth]{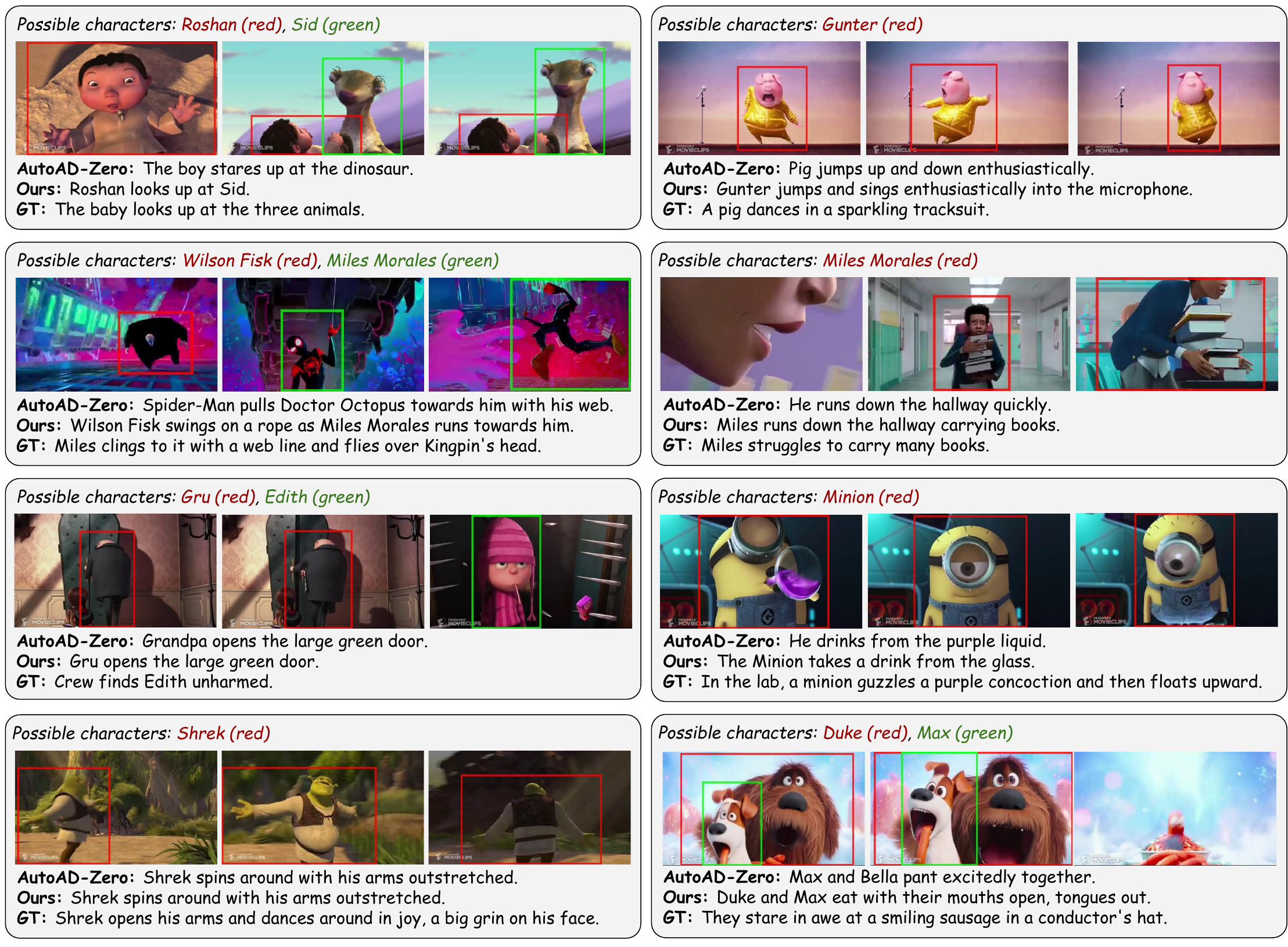}
    \caption{Visualisation of Generated Audio Descriptions in the CMD-AM Dataset. The examples are from \textit{Ice Age}, \textit{Sing}, \textit{Spider Man: Into the Spider-Verse}, \textit{Despicable Me}, \textit{Shrek}, and \textit{The Secret Life of Pets}.}
    \label{fig:ad_cmdam}
\end{figure*}

We provide more qualitative examples of our generated ADs for the CMD-AM dataset, illustrated in Fig.\ \ref{fig:ad_cmdam}.

\begin{figure*}[ht]
    \centering
    \includegraphics[width=0.95\linewidth]{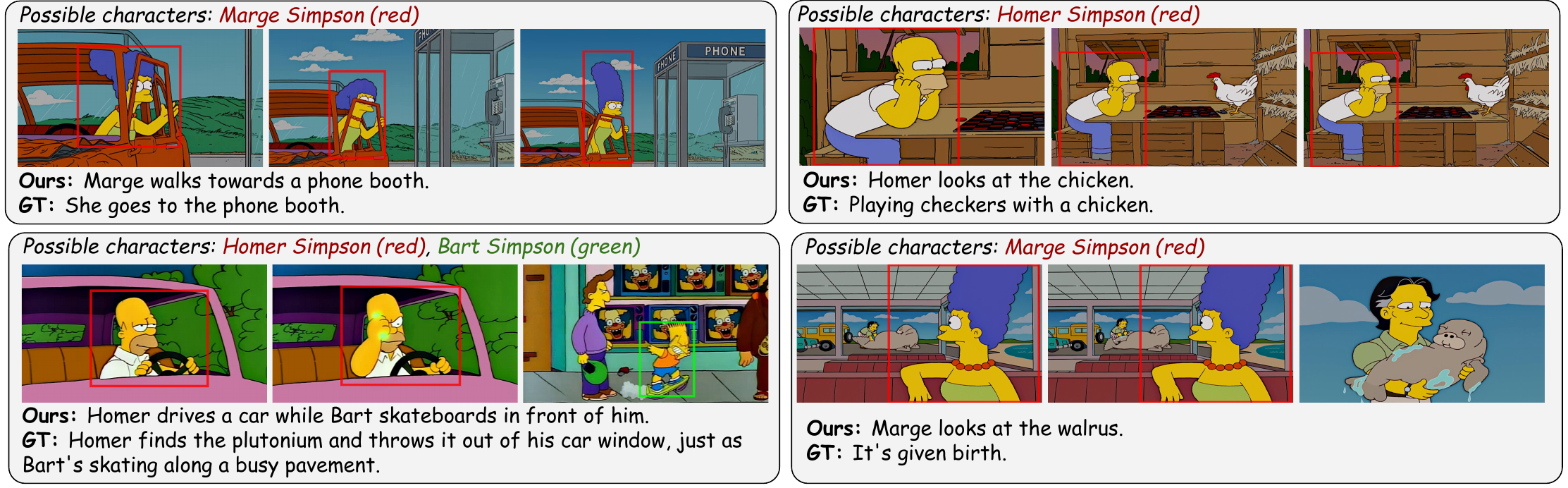}
    \caption{Visualisation of Generated Audio Descriptions for \textit{The Simpsons} Episodes.}
    \label{fig:ad_simpsons}
\end{figure*}

Moreover, we apply our AD generation method to other forms of animated media, specifically, \textit{The Simpsons}, as shown in Fig.\ \ref{fig:ad_simpsons}, demonstrating that our approach generalises well to different styles of animated content.

\subsection{Qualitative Results of Character-Aware Subtitles}
\begin{figure*}[ht]
    \centering
    \includegraphics[width=0.95\linewidth]{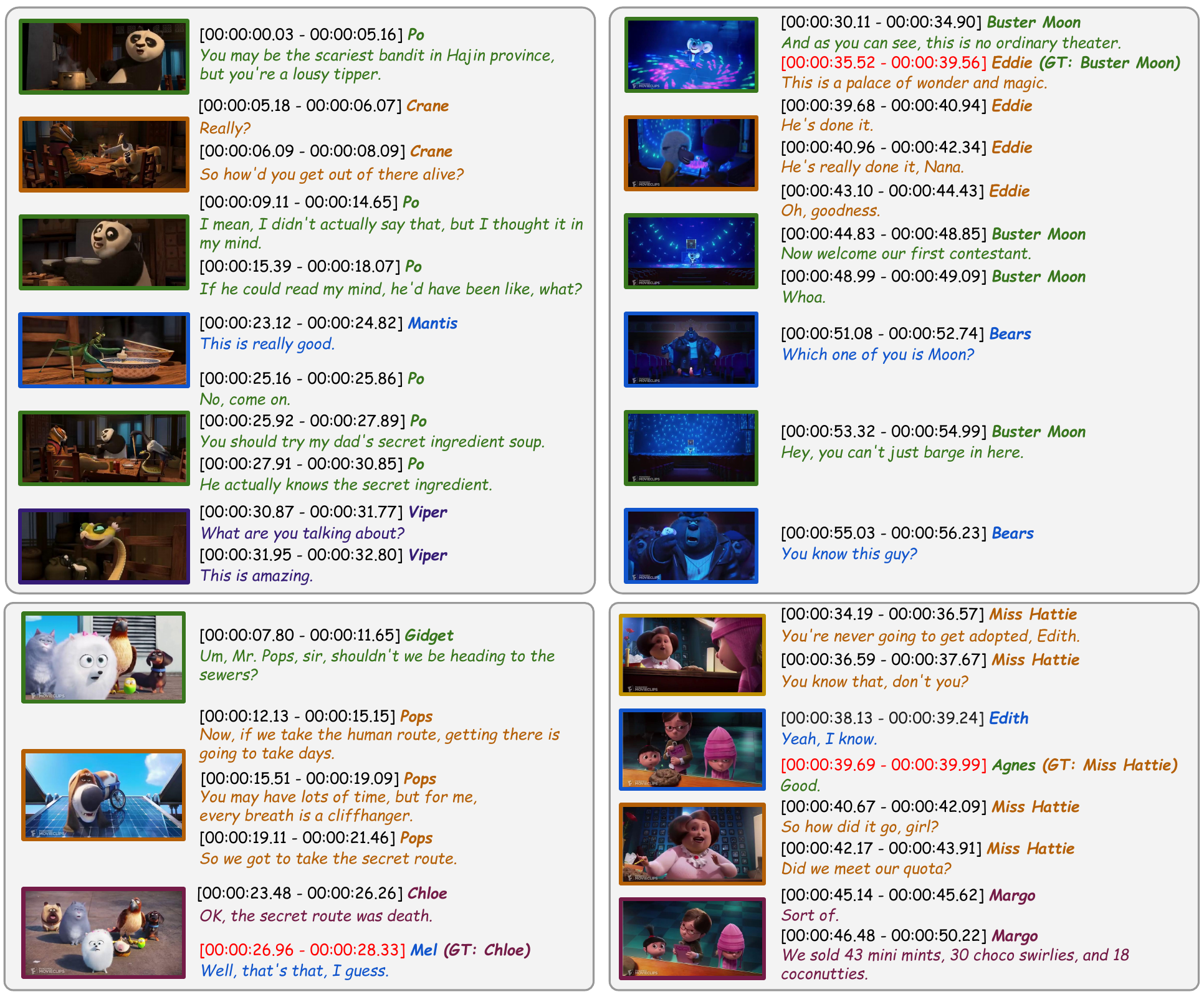}
    \caption{Visualisation of Generated Character-Aware Subtitles for Segments from \textit{Kung Fu Panda}, \textit{Sing}, \textit{The Secret Life of Pets}, and \textit{Despicable Me}. Timestamps highlighted in red denote incorrect audio identification.}
    \label{fig:subtitles_supp}
\end{figure*}

We provide more qualitative examples of our generated character-aware subtitles in Fig.\ \ref{fig:subtitles_supp}. We highlight incorrect subtitle predictions by marking the corresponding timestamps in red. The remaining predicted character labels are correct.


\end{document}